\documentclass[journal,twoside,web]{ieeecolor}

\usepackage{generic}
\usepackage{cite}
\usepackage{amsmath,amssymb,amsfonts,bm}
\usepackage{graphicx}
\usepackage{booktabs,tabularx,array,makecell}
\usepackage{placeins}
\usepackage{adjustbox}
\usepackage{rotating}
\usepackage{hyperref}
\hypersetup{hidelinks}
\usepackage{textcomp}

\newcommand{\method}{PFedHASNet}
\let\citep\cite

\newcommand{\KBSColumnTableFit}[1]{{%
  \footnotesize\setlength{\tabcolsep}{2.6pt}%
  \renewcommand{\arraystretch}{1.12}%
  \begin{adjustbox}{max width=\columnwidth}#1\end{adjustbox}%
}}

\begin{document}
\markboth{IEEE Journal of Biomedical and Health Informatics}
{Li \MakeLowercase{\textit{et al.}}: PFedHASNet}

\title{PFedHASNet: A Personalized Federated Hybrid Analog--Spike Network for Multi-Subject Motor Imagery EEG Decoding}

\author{%
\makebox[\textwidth][c]{%
\resizebox{0.98\textwidth}{!}{%
Peijia Li, Ruidong Ma, Xianghong Lin, Xiangwen Wang,
Guoyong Shi, Shuai Huang, and Hongqi Li%
}}%
\thanks{This work was supported in part by the
National Natural Science Foundation of China
under Grant Nos. 62266040 and 62666061.
This work has been submitted to the IEEE for possible
publication. Copyright may be transferred without
notice, after which this version may no longer be
accessible.}%
\thanks{Peijia Li and Ruidong Ma contributed equally
to this work.}%
\thanks{Corresponding author: Xianghong Lin
(e-mail: linxh@nwnu.edu.cn).}%
\thanks{Peijia Li, Ruidong Ma, Xianghong Lin,
Xiangwen Wang, and Guoyong Shi are with the
College of Artificial Intelligence and Computer Science,
Northwest Normal University, Lanzhou 730070, China
(e-mail: 202521162095@nwnu.edu.cn).}%
\thanks{Shuai Huang is with the Shanghai Key Lab
of Intelligent Information Processing, College of Computer Science and Artificial Intelligence, Fudan University, Shanghai 200438, China}%
\thanks{Hongqi Li is with the School of Software,
Northwestern Polytechnical University,
Xi'an, 710072, China.}}

\maketitle

\begin{abstract}
Motor imagery electroencephalography (MI-EEG)
data are often collected and stored separately
across subjects or institutions, and EEG
acquisition and annotation remain costly.
Training models solely on individual subject
data limits the use of information from other
subjects, whereas centralizing raw EEG
recordings raises privacy and data governance
concerns. Federated learning (FL) enables
collaborative model training across distributed
clients without sharing raw EEG data, providing
a practical approach to leveraging information
from multiple subjects. Accordingly, we propose
PFedHASNet, a personalized federated hybrid
analog--spike network for MI-EEG decoding.
The shared encoder integrates continuous-valued
temporal features, leaky integrate-and-fire (LIF)
spike responses, and channel-wise log-variance
statistics through learnable fusion weights.
Client-private residual adapters and classifiers
perform subject-specific feature refinement
and personalized prediction, respectively.
Experiments on held-out trials from participating
subjects across four public MI-EEG datasets
show that PFedHASNet achieves the highest mean
client macro-F1 among 18 model configurations
in cross-model comparisons, while maintaining
competitive decoding performance with a compact
client-side model in comparisons using the same
backbone. Ablation studies further examine the
contributions of hybrid representations and
personalization components to decoding performance.
\end{abstract}


\section{Introduction}
Motor imagery electroencephalography (MI-EEG) decodes
changes in sensorimotor rhythms to infer control intentions
and is widely studied in noninvasive brain--computer
interfaces (BCIs). Deep networks learn temporal, spatial,
and spectral features from EEG
signals~\citep{lawhern2018eegnet,mane2021fbcnet,song2023eegconformer}.
Studies have investigated inter-subject variability through
feature disentanglement, multi-source domain generalization,
and meta-learning~\citep{han2024noisefactorized,zhong2025eegdg,liu2026dameta}.
Unsupervised self-training has also been explored for
cross-domain MI decoding~\citep{chen2025sstda}.
However, signal amplitude, noise structure, spatial patterns,
and class boundaries vary substantially among
subjects~\citep{kwon2020subjectindependent,weng2026statemamba},
making it difficult for a single shared representation
to accommodate all subjects without personalization.

Beyond inter-subject variability, high-quality EEG
acquisition and annotation are costly, and data collected
by different subjects or institutions are commonly stored
separately. Centralized training requires pooling raw
signals, raising practical and data-governance concerns,
while training on an individual subject's data limits
access to patterns observed in other subjects.
Federated learning (FL) supports collaborative training
through model updates while keeping raw EEG local.
A client may represent a subject, device, or institution;
here, each client represents one subject.
In this setting, FedAvg aggregates locally trained
parameters according to client sample
sizes~\citep{mcmahan2017fedavg}, while FedProx introduces
a proximal regularizer to reduce drift under
heterogeneous data~\citep{li2020fedprox}.

A common aggregated model does not necessarily preserve
subject-specific features and decision boundaries.
Personalized federated learning (PFL) addresses this
limitation through shared--private parameter partitions
or coordinated global and local models.
FedPer and FedRep share representation parameters while
retaining client-private prediction
components~\citep{arivazhagan2019fedper,collins2021fedrep},
whereas Ditto learns separate personalized models under
regularization~\citep{li2021ditto}.
These methods provide different strategies for combining
shared representation learning with client-specific
prediction.

\begin{figure*}[!t]
  \centering
  \includegraphics[width=0.96\textwidth,keepaspectratio]{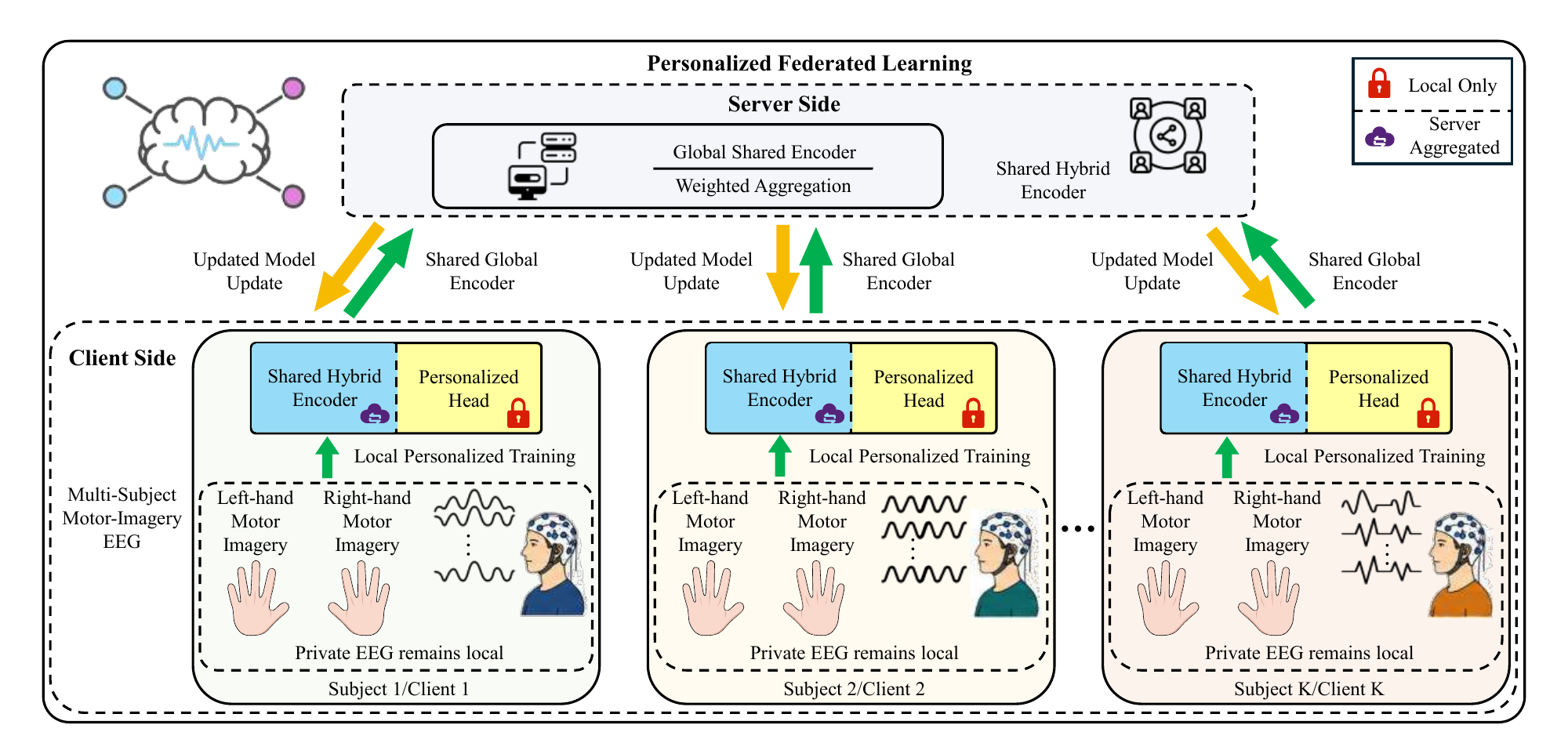}
  \caption{Personalized federated framework of \method. The server aggregates the shared encoder, while each client retains local EEG, a private residual adapter, and a private classifier for personalized local training.}
  \label{fig:framework}
\end{figure*}

Personalized distributed EEG learning has also been
studied for emotion recognition and cognitive-state
identification~\citep{qiu2025fedkdc,ouyang2026p3dl}.
These EEG studies address different tasks from MI decoding.
For MI-EEG, the shared encoder needs to represent temporal
fluctuations and channel-wise signal statistics while
supporting subject-specific feature correction and prediction.
The shared--private partition must also account for
client-held model size and repeated communication,
motivating compact private modules alongside a shared
EEG representation.

Spiking neural networks (SNNs) provide a pathway for
modeling temporal responses through membrane-potential
integration, threshold-triggered firing, and reset.
Supervised learning based on spike-train inner products
has been investigated for multilayer
SNNs~\citep{lin2017multistip}, while spike optimization
mechanisms have been developed to improve learning
efficiency~\citep{hu2022spikeoptimization}.
Spiking models have been applied to MI-EEG
decoding~\citep{ma2026drcanet} and studied in federated
settings~\citep{venkatesha2021fedsnn}.
Their spike-rate responses offer a temporal feature
pathway that can be combined with continuous-valued
features and channel-wise log-variance statistics.
Federated EEG studies have also examined distributed
classification and heterogeneous
collaboration~\citep{ju2020federatedeeg,jia2024fedbs,liu2024fleeg}.
These studies motivate examining hybrid EEG representations
together with compact client-private adaptation in
multi-subject federated MI-EEG decoding.

To this end, we propose \method{}, a Personalized Federated
Hybrid Analog--Spike Network for multi-subject MI-EEG
decoding. The framework combines a shared multiscale
spatiotemporal encoder and the Hybrid Analog--Spike
Representation Module (HASR) with client-private residual
adapters and classifiers. The shared encoder supports
collaborative representation learning, while the private
modules perform subject-specific feature correction and
prediction. The main contributions are as follows:
\begin{itemize}
\item \textbf{We propose a lightweight, personalized
federated framework for EEG decoding of motor imagery
across multiple subjects.}
PFedHASNet combines shared representation learning
with subject-specific adaptation; it supports
collaborative learning among clients via a shared
encoder and utilizes client-private residual adapters
and classifiers to achieve feature refinement and
personalized prediction, respectively.

\item \textbf{We designed a hybrid analog--spike
representation module that fuses temporal features
with channel statistics.}
HASR fuses continuous-valued temporal summaries
extracted from spatio-temporal features, LIF pulse
firing rates, and channel-level logarithmic variance
statistics extracted directly from EEG inputs to
construct a shared representation for personalized
decoding via learnable fusion weights.

\item \textbf{We conducted multi-dataset evaluations
to analyze the roles of representation learning
and personalized design.}
On four public MI-EEG datasets, we evaluated the
framework's decoding performance through cross-model
comparisons, comparisons with backbone federated
strategies, component ablation, and client-level
analysis. We characterized the contributions of
each component and the differences in benefits
across participants, and examined the trade-offs
between performance and client-side model size.
\end{itemize}

\begin{figure*}[!t]
  \centering
  \includegraphics[width=0.97\textwidth,keepaspectratio]{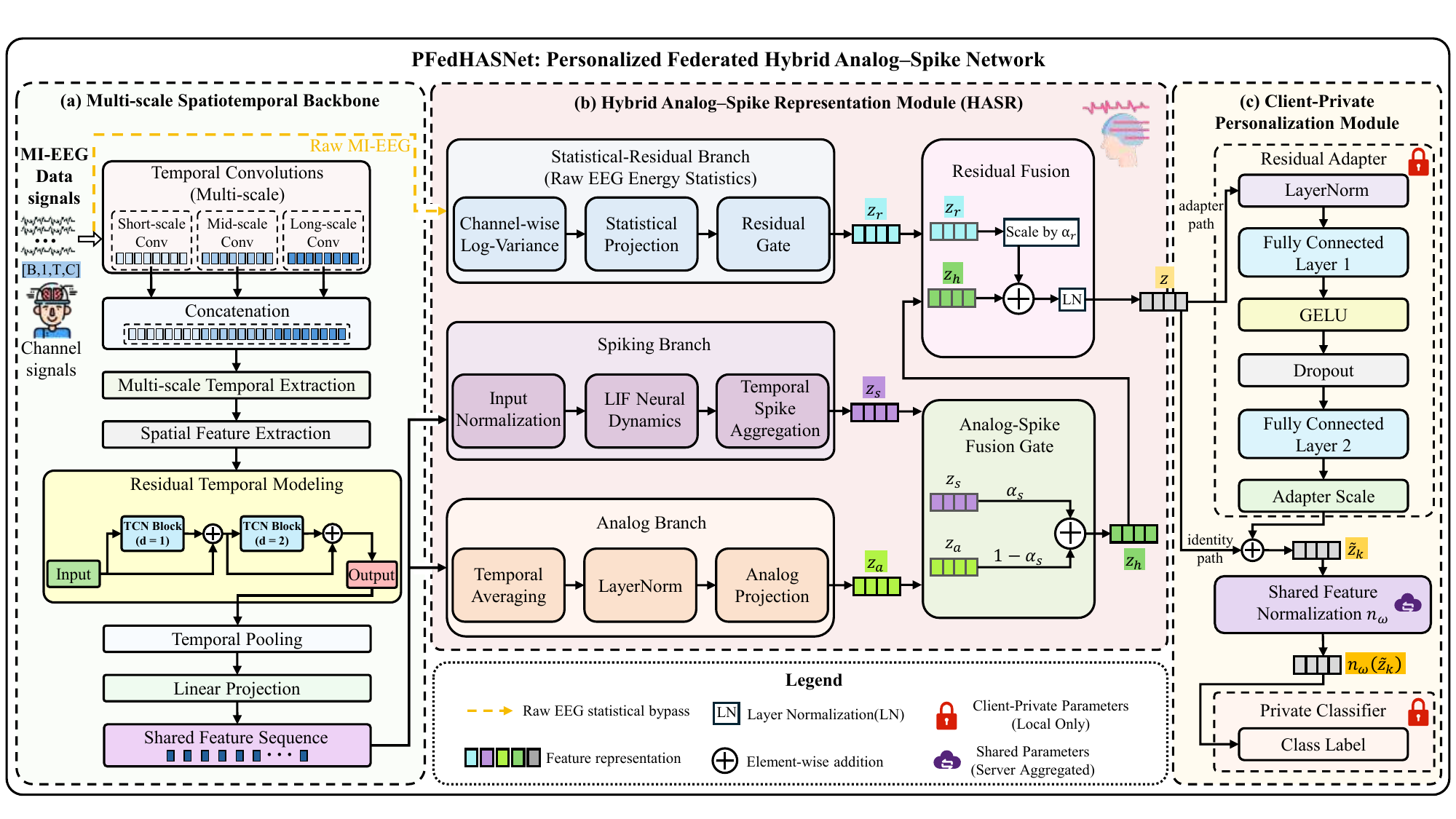}
  \caption{Architecture of \method. (a) Multi-scale spatiotemporal feature extraction backbone. (b) HASR, which integrates a continuous-valued summary, LIF spike rates, and channel-wise LogVar statistics. (c) Client-private residual adapter and private classifier.}
  \label{fig:model}
\end{figure*}

\section{Related Work}

\subsection{Motor Imagery EEG Decoding}

Deep MI-EEG models employ compact spatiotemporal
convolutions, spectral--spatial statistics, and temporal
convolution or attention for longer-range
dependencies~\citep{lawhern2018eegnet,schirrmeister2017deepconvnets,mane2021fbcnet,altaheri2023atcnet,song2023eegconformer}.
Recent work has explored multiscale spatial--temporal fusion,
lightweight residual convolution, and fine-grained
spatial--frequency--time features for MI decoding~\citep{jin2025mstfnet,li2025mfrcnet,liu2025fgsft}.
SST-DPN combines spatial--spectral attention, multiscale
variance pooling, and dual prototype
learning~\citep{han2025sstdpn}.
Context-based adaptation has been explored to address
cross-session variability~\citep{wang2025bsan}.
Those methods motivate richer EEG
representations, but do not determine how shared and
client-specific parameters should be organized during
federated training.

\subsection{Federated Learning for EEG}
Federated learning enables collaborative training across decentralized clients without pooling raw data. FedAvg aggregates locally updated model parameters according to client sample sizes~\citep{mcmahan2017fedavg}, while FedProx introduces a proximal term to reduce client drift under heterogeneous data distributions~\citep{li2020fedprox}. Both methods support collaborative optimization through global parameter aggregation. However, a unified model may have difficulty accommodating differences in signal distributions and decision boundaries across subjects, motivating PFL.

PFL methods address client heterogeneity through different parameter-sharing and optimization strategies. FedPer separates shared feature extraction layers from client-private prediction layers and jointly updates them during local training~\citep{arivazhagan2019fedper}. FedRep also learns shared representations and private classification heads but alternates their optimization during local training~\citep{collins2021fedrep}. Ditto maintains a global model alongside a personalized model for each client, using regularization to connect local optimization with the global model~\citep{li2021ditto}. These approaches provide different ways to balance shared knowledge and client-specific prediction, with distinct parameter allocations and local training procedures.

For heterogeneous EEG datasets, the Sandwich framework combines
federated and transfer learning with dataset-specific input
networks, a shared network, and individual classifiers~\citep{wei2025sandwich}.

For multi-subject MI-EEG decoding, \method{} extends shared--private personalization with an EEG-specific hybrid analog--spike encoder and a private residual adapter for feature correction. Local training jointly updates the encoder and private modules; the server aggregates shared parameters according to sample counts. This design combines hybrid EEG representation learning with subject-specific prediction.

Supervised SNN algorithms and evaluation practices
have been reviewed by Wang et al.~\citep{wang2020snnreview}.
Subsequent studies have investigated one-pass online
learning using gradient descent for multilayer
SNNs~\citep{lin2023onepass}
and joint synaptic weight--delay
learning~\citep{wang2026swdsl}.
For EEG applications, spiking networks have been
studied for MI decoding~\citep{ma2026drcanet}
and epilepsy detection~\citep{su2026stesnn};
federated SNN research has also examined
decentralized training~\citep{venkatesha2021fedsnn}.
In PFedHASNet, average LIF firing rates serve as one
feature pathway alongside continuous-valued temporal
features and channel-wise LogVar statistics.

\section{Method}

\subsection{Problem Formulation and Overall Framework}

Consider $K$ participating subjects, each acting as one client. Client $k$ holds $D_k=\{(x_i,y_i)\}_{i=1}^{n_k}$, where $x_i\in\mathbb{R}^{C\times T}$ denotes an EEG trial with $C$ channels and $T$ samples, and $y_i\in\{1,\ldots,M\}$ denotes its MI label. Collaborative training and evaluation use the same participating subjects, with disjoint training and test trials for each subject (participating-client personalization).

As illustrated in Figure~\ref{fig:framework}, \method{} combines a shared encoder with client-private personalization modules. In each communication round, the server distributes the shared parameters to all clients. Each client jointly updates its shared and private parameters using local EEG, but uploads only the shared parameters for server aggregation. The private residual adapter and classifier always remain local. This design preserves subject-specific feature correction and category decision-making capabilities while learning cross-subject shared representations.

As shown in Figure~\ref{fig:model}, \method{} consists of three components: (a) a multi-scale spatiotemporal feature extraction backbone that captures local spatiotemporal patterns and long-range temporal dependencies through multi-scale temporal convolution, spatial convolution, and a residual dilated TCN; (b) the Hybrid Analog--Spike Representation Module (HASR), which integrates a continuous-valued summary, LIF spike rates, and channel-wise LogVar statistics into a hybrid EEG representation; and (c) a client-private personalization module comprising a private residual adapter and classifier for subject-specific feature correction and decision boundaries.

Components (a) and (b) jointly form the shared encoder and participate in server aggregation. In component (c), the residual adapter and classifier are private, whereas the intervening feature normalization layer is shared and aggregated. For client $k$, the input EEG is first mapped to a hybrid representation by the shared encoder, corrected by the private residual adapter, normalized by the shared feature normalization layer, and then passed to the private classifier. The forward process is
\begin{equation}
P_k(y\mid x)=
\operatorname{softmax}\left(
h_{\psi_k}\left[
n_{\omega}\left(
a_{\phi_k}\left(f_{\theta,\omega}(x)\right)
\right)\right]\right),
\label{eq:prediction}
\end{equation}
where $f_{\theta,\omega}$ comprises the backbone and HASR, $a_{\phi_k}$
is the private residual correction, and $h_{\psi_k}$ is the private classifier.
The same shared LayerNorm $n_\omega$ is used inside HASR fusion and again
after the residual correction; these are two calls to one parameterized layer.
We define disjoint shared parameters $\Theta^s=\{\theta,\omega\}$, with
$\theta$ excluding this LayerNorm, and private parameters
$\Theta_k^p=\{\phi_k,\psi_k\}$.

Based on the above shared--private parameter partitioning, client $k$ minimizes the cross-entropy (CE) empirical risk using local data:
\begin{equation}
\mathcal{L}_k(\Theta^{s},\Theta_k^{p})
=
-\frac{1}{n_k}
\sum_{(x,y)\in D_k}
\log P_k(y\mid x).
\label{eq:local_risk}
\end{equation}
Federated training jointly optimizes the shared parameters through sample-size-weighted aggregation while retaining each client's private parameters. The overall objective is
\begin{equation}
\min_{\Theta^{s},\{\Theta_k^{p}\}_{k=1}^{K}}
\sum_{k=1}^{K}
\frac{n_k}{\sum_{j=1}^{K}n_j}
\mathcal{L}_k(\Theta^{s},\Theta_k^{p}).
\label{eq:federated_objective}
\end{equation}

\subsection{Multi-Scale Spatiotemporal Feature Extraction}

As illustrated in Figure~\ref{fig:model}(a), the shared
spatiotemporal backbone transforms each preprocessed EEG trial
$X\in\mathbb{R}^{C\times T}$ into a temporally ordered feature
sequence. The input is processed by parallel temporal
convolution branches with different receptive fields, capturing
short-term signal fluctuations and rhythmic patterns over
multiple temporal scales. Their outputs are concatenated to
integrate temporal information. Subsequent
spatial convolution models the relationships among EEG
channels, allowing the network to extract spatially
discriminative patterns from the multi-scale features.

The resulting features are further processed by a residual
temporal convolutional network (TCN) comprising two dilated
blocks with dilation factors of 1 and 2, respectively.
Dilated convolutions expand the temporal receptive field
to capture contextual dependencies beyond the initial
convolutional stages, while residual connections preserve
preceding representations and facilitate feature propagation.
Temporal pooling then reduces the sequence length, followed
by linear projection and GELU activation to obtain a
compact feature representation.

The backbone produces a feature sequence
$H=[h_1,\ldots,h_{T'}]\in\mathbb{R}^{T'\times d}$,
where $T'$ denotes the reduced temporal length and
$d=64$ is the feature dimension. This sequence provides
the shared input to the analog and spiking branches of
HASR, as shown in Figure~\ref{fig:model}(b).
In parallel, the preprocessed EEG input is directly
routed to the channel-wise LogVar branch, preserving
signal-level variance information independently of the
learned convolutional features. This design enables
HASR to combine learned spatiotemporal representations
with complementary channel-wise statistics.

\subsection{Hybrid Analog--Spike Representation Module}

As shown in Figure~\ref{fig:model}(b), HASR follows the multi-scale spatiotemporal feature extraction module and contains three parallel branches. First, the continuous-valued branch temporally averages and normalizes $H$ to preserve continuous features from the spatiotemporal backbone. Second, after feature normalization, the LIF spike-rate pathway performs membrane-potential integration, threshold-triggered firing, and reset along the temporal dimension, and forms a spike-based representation from the mean spike rate. Third, the channel-wise LogVar branch extracts the log-variance of each channel from the preprocessed EEG input to summarize relative channel-wise amplitude variation. The branch outputs are
\begin{equation}
\left\{
\begin{aligned}
z_a &= \operatorname{LN}_a\left(T'^{-1}\sum_{t=1}^{T'}h_t\right),\\
s_t &= \operatorname{LIF}\!\left(\gamma\operatorname{LN}_s(h_t),v_{t-1}\right),\\
z_s &= T'^{-1}\sum_{t=1}^{T'}s_t,\\
z_r &= \mathcal P_r\!\left(\operatorname{LN}_C
\left[\log\!\left(\operatorname{Var}_t(X)+10^{-4}\right)\right]\right).
\end{aligned}
\right.
\label{eq:hasr_branches}
\end{equation}
All three outputs have dimension $d=64$; $H\in\mathbb R^{T'\times64}$,
with $T'=250$ for 1000 input samples. The analog branch averages time
before LayerNorm and has no additional projection. The LogVar input $X$
is the preprocessed EEG, not the backbone feature sequence.
$\operatorname{Var}_t$ is population variance (denominator $T$);
$\operatorname{LN}_C$ normalizes its $C$ channel values.
$\mathcal P_r$ applies a $C$-to-64 linear map, GELU, and dropout with
probability 0.1.

The positive scalar input gain is
$\gamma=\operatorname{softplus}(g_{\rm in})$.
For each trial and feature coordinate, the implemented
LIF update is
\begin{equation}
\label{eq:lif-update}
\left\{
\begin{aligned}
u_t &= e^{-1/\tau}v_{t-1}
       +\gamma\operatorname{LN}_s(h_t),\\
s_t &= \mathbf{1}[u_t>\vartheta],\\
v_t &= u_t(1-s_t),\qquad v_0=0.
\end{aligned}
\right.
\end{equation}
The fixed constants are $\tau=2$ and
$\vartheta=0.35$.
The membrane is initialized on every forward call
and is not carried between trials.
The reset is hard, and $z_s$ is the mean binary
firing rate over time.
Backpropagation replaces the threshold derivative
at $q=u_t-\vartheta$ by $(1+|q|)^{-2}$;
the reset is not detached.
The LayerNorms in the analog and spike branches
are not tied.

The three branches are fused through learnable gating:
\begin{equation}
z=n_\omega\left[
(1-\alpha_s)z_a+\alpha_s z_s+\alpha_r z_r
\right],
\label{eq:hasr_fusion}
\end{equation}
where $\alpha_s=\sigma(g_s)$ and $\alpha_r=\sigma(g_r)$ are
input-independent trainable scalar gates. The first interpolates
between analog and spike representations; the second independently
scales the LogVar residual. The raw scalar initializations are
$g_s=0.55$, $g_r=0.30$, and $g_{\rm in}=1.2$, giving effective
values approximately 0.634, 0.574, and 1.463, respectively.
These parameters and $n_\omega$ belong to the shared state
and are aggregated.
\subsection{Client-Private Personalization Module}

As shown in Figure~\ref{fig:model}(c), each client
retains a private residual adapter and classifier.
Given the shared representation $z$, the adapter
performs subject-specific feature correction as
\begin{equation}
a_{\phi_k}(z)
= z + \sigma(\beta_k)\mathcal{R}_{\phi_k}(z),
\label{eq:personalization}
\end{equation}
where $\mathcal{R}_{\phi_k}$ denotes a two-layer
bottleneck network comprising private LayerNorm,
a $64\to32$ linear projection, GELU, dropout
($p=0.1$), and a $32\to64$ linear projection.
The trainable scalar $\beta_k$, initialized to
$0.2$, controls the residual contribution through
a sigmoid function.

The corrected representation is passed through
the shared normalization layer $n_\omega$ and
the private classifier $h_{\psi_k}$, as defined
in Eq.~\eqref{eq:prediction}. The classifier
comprises private LayerNorm, a $64\to64$ linear
layer, GELU, dropout ($p=0.2$), and a final
classification layer. All adapter and classifier
parameters remain client-private across rounds
and are excluded from server aggregation.

\subsection{Personalized Federated Optimization}

At communication round $t$, the server distributes shared parameters $\Theta^{s,t}$ to all clients. Client $k$ loads its local private parameters $\Theta_k^{p,t}$ and jointly optimizes the shared and private components:
\begin{equation}
\mathcal{L}_{k}^{\mathrm{local}}
=
\mathcal{L}_{k}^{\mathrm{CE}}
+\frac{\mu}{2}
\left\|
\Theta_k^{s}-\Theta^{s,t}
\right\|_2^2,
\label{eq:local_objective}
\end{equation}
where the proximal term penalizes deviations of the local shared parameters from the shared state downloaded at the beginning of the round. For the formal PFedHASNet configuration, $\mu=0.0006$; the matched factorial ablation and FedAvg use $\mu=0$, and FedProx uses $\mu=0.001$. Equation~\eqref{eq:federated_objective} states the aggregated empirical-risk objective, while Eq.~\eqref{eq:local_objective} specifies the optional proximal regularization used in each client's updates. AdamW applies decoupled weight decay, and its optimizer is reinitialized for each client update.

After local training, the shared parameters are aggregated according to client sample sizes, while the private parameters are retained for the next round:
\begin{equation}
\left\{
\begin{aligned}
\Theta^{s,t+1}
&=\sum_{k=1}^{K}
\frac{n_k}{\sum_{j=1}^{K}n_j}
\Theta_k^{s,t+1},\\
\Theta_k^{p,t+1}
&\leftarrow
\Theta_{k,\mathrm{local}}^{p,t+1}.
\end{aligned}
\right.
\label{eq:federated_optimization}
\end{equation}

Aggregation weights are proportional to training-set sizes and all
clients participate. The implementation averages floating-point
state entries and copies non-floating entries from the last client;
PFedHASNet has no registered buffers, so its transmitted state consists
of parameters. The shared LayerNorm is included and private states are
retained locally.

Each client uploads only the shared parameters of the
spatiotemporal feature extractor, HASR, and shared normalization
layer. The private residual adapter and classifier do not
participate in aggregation. Raw EEG and sample-level intermediate
features remain local.

\section{Experiments}

\subsection{Datasets and Experimental Settings}

We evaluate all methods on four public motor imagery EEG (MI-EEG)
datasets: BCI Competition IV Dataset 2b (BNCI 2014-004)
\citep{leeb2007virtual}, BCI Competition IV Dataset 1
\citep{blankertz2007noninvasive}, BCI Competition IV Dataset 2a
(BNCI 2014-001) \citep{brunner2007spatial}, and Weibo2014
\citep{yi2014weibo}. Each subject constitutes one federated client,
with training on local training trials and evaluation on held-out test
trials from the same subject. Signals are linearly detrended,
band-pass filtered at 8--30 Hz, and resampled to 250 Hz; each 4-s
trial gives an input $x\in\mathbb{R}^{C_d\times1000}$, with $C_d$
channels for dataset $d$ (Table~\ref{tab:datasets-settings}).

Training, validation, and test trials are stored in separate,
prepartitioned per-subject arrays. The recorded total
training/validation/test trial counts are 3,058/622/2,840 for M1,
1,162/238/1,125 for M2, 2,160/432/2,592 for M3, and
1,956/414/2,370 for M4. The same predefined arrays are selected
for all three random seeds; no new seed-specific 80/20 split is
created. These recorded counts do not independently authenticate
trial-level split identities or session separation.

Federated configurations run for 100 rounds with full client
participation and two local epochs per round by default; FedRep
and Ditto follow the method-specific schedules in
Table~\ref{tab:same-backbone}. The recorded optimization settings
are AdamW, learning rate $10^{-3}$, weight decay $10^{-4}$,
batch size 64, and maximum gradient norm 1.0. The default dropout
is 0.2, with module-specific rates stated in Section~III.
Endpoint results use the prespecified round 100, without early
stopping or test-based checkpoint selection. Round-wise test
measurements are used for retrospective training-dynamics
analysis. Central and Local are evaluated at epoch 200.

\begin{table}[t]
  \centering
  \caption{Dataset characteristics and input dimensions. Each trial contains 1,000 samples, corresponding to 4~s at 250~Hz after preprocessing.}
  \label{tab:datasets-settings}
  \small
  \setlength{\tabcolsep}{2.7pt}
  \renewcommand{\arraystretch}{1.10}
  \begin{tabularx}{\columnwidth}{@{}>{\raggedright\arraybackslash}Xccc@{}}
    \toprule
    Dataset (ID) & Clients & Channels & Classes \\
    \midrule
    BCI Competition IV Dataset 2b (M1) & 9 & 3 & 2 \\
    BCI Competition IV Dataset 1 (M2) & 7 & 59 & 2 \\
    BCI Competition IV Dataset 2a (M3) & 9 & 22 & 4 \\
    Weibo2014 (M4) & 10 & 60 & 6 \\
    \bottomrule
  \end{tabularx}
\end{table}

For each dataset and model configuration, the endpoint performance is summarized across three independent random seeds (42, 2026, and 3407) and reported as the mean $\pm$ sample standard deviation (SD). For each seed, the primary metric is calculated by first computing macro-F1 over the classes of each participating client and then averaging the client-level scores with equal weight, as defined in Eq.~\eqref{eq:mean-client-f1}. The resulting dataset-level scores are subsequently averaged across seeds. The sample SD is calculated from the three seed-level scores using a denominator of $3-1$ and reflects variation across random seeds, rather than variation among clients or a confidence interval. Accuracy, macro-precision, macro-recall, and test cross-entropy are summarized using the same client-then-seed averaging procedure. Cross-entropy loss is reported on its original numerical scale and is excluded from performance ranking.

To summarize performance across datasets, Average is calculated as the arithmetic mean of the four dataset-level mean scores (M1--M4), assigning equal weight to each dataset regardless of its number of clients or test trials. Mean Rank is obtained by ranking the compared model configurations according to their mean macro-F1 within each dataset and averaging these ranks across the four datasets, with rank 1 indicating the highest score. These complementary summaries report the overall predictive performance and the relative ranking of each configuration across the evaluated datasets.

\begin{table*}[!t]
  \centering
  \caption{Cross-model mean-client macro-F1
  (\%; mean $\pm$ sample SD across three seeds)
  at round 100. Average and Mean Rank summarize
  four datasets; bold and underlining mark the
  best and second-best means.}
  \label{tab:all-federated-models}

  \small
  \setlength{\tabcolsep}{3pt}
  \renewcommand{\arraystretch}{1.12}

  \begin{tabular*}{\textwidth}{@{}l@{\extracolsep{\fill}}cccccc@{}}
  \toprule
  Configuration & M1 & M2 & M3 & M4 & Average & Mean Rank \\
  \midrule

  ShallowConvNet~\citep{schirrmeister2017deepconvnets}
  & 71.89$\pm$1.40
  & 63.99$\pm$1.82
  & \underline{44.54$\pm$1.44}
  & \underline{25.76$\pm$1.10}
  & \underline{51.55}
  & 5.00 \\

  ADFCNN~\citep{tao2024adfcnn}
  & 75.12$\pm$0.39
  & \underline{65.70$\pm$2.63}
  & 42.90$\pm$1.23
  & 19.64$\pm$0.32
  & 50.84
  & 4.00 \\

  DSAINet~\citep{ma2026dsainet}
  & 72.77$\pm$1.54
  & 56.83$\pm$0.58
  & 42.58$\pm$1.24
  & 13.70$\pm$2.01
  & 46.47
  & 9.75 \\

  TMSA-Net~\citep{zhao2025tmsanet}
  & 73.67$\pm$1.15
  & 59.86$\pm$2.59
  & 39.73$\pm$1.74
  & 12.24$\pm$4.35
  & 46.38
  & 9.25 \\

  EEGNet~\citep{lawhern2018eegnet}
  & 73.85$\pm$0.23
  & 58.57$\pm$0.61
  & 37.25$\pm$0.46
  & 11.83$\pm$0.88
  & 45.37
  & 10.75 \\

  DeepConvNet~\citep{schirrmeister2017deepconvnets}
  & 71.72$\pm$0.76
  & 47.23$\pm$2.98
  & 38.19$\pm$0.88
  & 19.06$\pm$0.45
  & 44.05
  & 12.00 \\

  LMDA-Net~\citep{miao2023lmdanet}
  & 72.89$\pm$1.48
  & 49.05$\pm$5.54
  & 32.63$\pm$0.46
  & 14.79$\pm$2.91
  & 42.34
  & 13.50 \\

  EEG-Conformer~\citep{song2023eegconformer}
  & 73.91$\pm$0.47
  & 57.26$\pm$1.15
  & 41.16$\pm$0.87
  & 21.30$\pm$2.19
  & 48.40
  & 6.75 \\

  MSCFormer~\citep{zhao2025mscformer}
  & \underline{75.94$\pm$0.61}
  & 59.84$\pm$3.42
  & 36.98$\pm$0.49
  & 18.05$\pm$0.80
  & 47.70
  & 7.00 \\

  DBConformer~\citep{wang2026dbconformer}
  & 74.12$\pm$0.98
  & 56.54$\pm$3.73
  & 34.44$\pm$0.79
  & 20.57$\pm$1.81
  & 46.42
  & 9.25 \\

  CTNet~\citep{zhao2024ctnet}
  & 74.25$\pm$1.19
  & 52.92$\pm$1.90
  & 41.25$\pm$1.01
  & 13.69$\pm$0.60
  & 45.53
  & 9.50 \\

  MSVTNet~\citep{liu2024msvtnet}
  & 75.33$\pm$1.17
  & 55.49$\pm$1.56
  & 35.19$\pm$0.64
  & 9.55$\pm$1.79
  & 43.89
  & 11.50 \\

  EEG-Deformer~\citep{ding2025eegdeformer}
  & 71.26$\pm$1.03
  & 43.29$\pm$0.58
  & 36.42$\pm$0.48
  & 19.07$\pm$3.29
  & 42.51
  & 13.00 \\

  MGFormer~\citep{rahman2025mgformer}
  & 68.48$\pm$1.22
  & 52.23$\pm$0.81
  & 32.18$\pm$0.84
  & 8.05$\pm$1.43
  & 40.24
  & 16.25 \\

  FANN~\citep{yang2026ibcis}
  & 52.72$\pm$0.29
  & 52.38$\pm$0.46
  & 30.13$\pm$0.12
  & 16.94$\pm$0.80
  & 38.04
  & 14.75 \\

  iBCIs-SNN~\citep{yang2026ibcis}
  & 45.09$\pm$10.45
  & 58.57$\pm$2.77
  & 25.55$\pm$3.18
  & 17.55$\pm$0.48
  & 36.69
  & 13.25 \\

  Hybrid-FedAvg
  & 75.88$\pm$0.65
  & 63.87$\pm$1.25
  & 39.16$\pm$0.49
  & 22.85$\pm$1.46
  & 50.44
  & 4.50 \\

  \textbf{\method{} (Ours)}
  & \textbf{76.20$\pm$0.48}
  & \textbf{66.10$\pm$7.85}
  & \textbf{47.40$\pm$1.32}
  & \textbf{27.65$\pm$0.61}
  & \textbf{54.34}
  & \textbf{1.00} \\

  \bottomrule
  \end{tabular*}

  \par\smallskip
  \parbox{\linewidth}{\footnotesize
  Average and Mean Rank summarize performance across the four datasets.}

\end{table*}

We use mean-client macro-F1 as the primary evaluation metric:
\begin{equation}
F_{d,s}=\frac{1}{K_d}\sum_{k=1}^{K_d}
\left(\frac{1}{M_d}\sum_{c=1}^{M_d}
\frac{2P_{d,k,c,s}R_{d,k,c,s}}{P_{d,k,c,s}+R_{d,k,c,s}}\right),
\label{eq:mean-client-f1}
\end{equation}
where $P_{d,k,c,s}$ and $R_{d,k,c,s}$ denote class-wise precision and
recall for client $k$, dataset $d$, and seed $s$; $K_d$ and $M_d$
are the numbers of clients and classes. A class-wise F1 is zero when
its precision and recall sum to zero. Clients are equally weighted
in evaluation, whereas shared parameters are aggregated using client
training-set sizes. $F_{d,s}$ is fraction-valued: scores are multiplied
by 100 for percentages, and differences between such scores are in pp.

For each endpoint metric, we report the mean and sample SD across
seeds 42, 2026, and 3407; the SD uses denominator $3-1$ and is neither
a client-level SD nor a confidence interval. Average is the unweighted
mean of the four dataset means; Mean Rank averages within-dataset
ranks, with rank 1 indicating the highest mean. Accuracy,
macro-precision, macro-recall, and test cross-entropy follow the same
client-then-seed averaging. Loss is unscaled and is not used for the
primary ranking. The aggregation diagnostics in
Section~\ref{sec:training-dynamics}
instead summarize temporally dependent paired seed--round changes.
Overall configuration comparisons and matched module ablations are
interpreted separately under their stated settings.

\subsection{Unified Cross-Model Comparison}

Table~\ref{tab:all-federated-models} compares 18 representative
model configurations spanning convolutional and temporal-convolutional
networks, Transformer/Conformer architectures, SNNs, and hybrid
continuous-valued and spiking networks. The configurations follow
the preprocessing and basic training settings described above.
Except for PFedHASNet, all models are trained with FedAvg.
Hybrid-FedAvg uses our hybrid continuous-valued and spiking backbone
without client personalization.

PFedHASNet attains the highest mean score on M1–M4, with an average of 54.34\% and a mean rank of 1.00. Relative to the second-highest mean on each dataset, PFedHASNet improves by 0.26, 0.40, 2.86, and 1.89 percentage points (pp) on M1–M4, respectively, with larger margins on M3 and M4. Its four-dataset average exceeds that of ShallowConvNet, the second-best by average score, by 2.79 pp.

Hybrid-FedAvg achieves a four-dataset average macro-F1 of
50.44\% and a mean rank of 4.50, compared with 54.34\%
and 1.00 for PFedHASNet. Across the three random seeds,
PFedHASNet ranks first among the evaluated configurations
on M3 and M4 in every run and on M1 in two runs. On M2,
PFedHASNet obtains 74.97\%, 60.05\%, and 63.29\% for
seeds 42, 2026, and 3407, respectively, compared with
68.40\%, 63.14\%, and 65.57\% for ADFCNN. Its M2
performance varies across seeds, with a mean macro-F1
of $66.10\pm7.85$\%.

Accuracy and macro-F1 yield different model rankings on
some datasets. On M3, PFedHASNet and ShallowConvNet
achieve similar accuracy (48.61\% and 48.55\%), while
PFedHASNet obtains a higher macro-F1 (47.40\% versus
44.54\%). On M2, PFedHASNet achieves a macro-F1 of
66.10\% and an accuracy of 69.05\%, compared with
65.70\% and 69.64\% for ADFCNN, respectively. On M4,
PFedHASNet achieves the highest accuracy (30.25\%),
macro-precision (33.62\%), macro-recall (30.25\%),
and macro-F1 (27.65\%) among the 18 configurations.
These results demonstrate the effectiveness of PFedHASNet
in multi-class MI-EEG decoding, with notable macro-F1
improvements on the four-class M3 and six-class M4 datasets.

\begin{table*}[!t]
  \centering
  \caption{Same-backbone comparison:
  (a) mean-client macro-averaged F1 score
  (\%, mean $\pm$ sample SD across three seeds);
  (b) training protocols and nominal costs.
  Federated methods A--G use 100 rounds;
  Central and Local use 200 epochs.}
  \label{tab:same-backbone}

  \small
  \setlength{\tabcolsep}{3pt}
  \renewcommand{\arraystretch}{1.10}

  \noindent\textit{(a) Performance: mean-client macro-averaged F1 score
  (\%; mean $\pm$ sample SD)}
  \par\smallskip

  \begin{tabular*}{\textwidth}
    {@{\extracolsep{\fill}}clccccc@{}}
  \toprule
  Code & Method & M1 & M2 & M3 & M4 & Average \\
  \midrule

  \multicolumn{7}{@{}l@{}}{
    \textit{Non-federated references
    (excluded from A--G ranking)}} \\
  \addlinespace[1pt]

  -- & Central
  & 74.33$\pm$0.98
  & 72.18$\pm$0.37
  & 48.76$\pm$0.84
  & 30.92$\pm$0.77
  & 56.55 \\

  -- & Local
  & 70.72$\pm$2.08
  & 68.01$\pm$4.07
  & 50.10$\pm$1.68
  & 30.75$\pm$0.83
  & 54.89 \\

  \midrule

  \multicolumn{7}{@{}l@{}}{
    \textit{Federated configurations A--G
    (same hybrid continuous-valued and spiking backbone)}} \\
  \addlinespace[1pt]

  A & FedAvg
  & \underline{75.88$\pm$0.65}
  & 63.87$\pm$1.25
  & 39.16$\pm$0.49
  & 22.85$\pm$1.46
  & 50.44 \\

  B & FedProx
  & 75.86$\pm$0.61
  & 64.04$\pm$1.36
  & 39.38$\pm$0.53
  & 22.82$\pm$1.45
  & 50.52 \\

  C & FedPer
  & 75.45$\pm$0.76
  & 64.38$\pm$3.03
  & 48.58$\pm$0.51
  & 27.85$\pm$1.03
  & 54.07 \\

  D & FedRep
  & 72.31$\pm$0.84
  & 65.27$\pm$3.12
  & \textbf{48.95$\pm$2.47}
  & \underline{29.59$\pm$1.23}
  & 54.03 \\

  E & \makecell[cl]{\method{}\\(without adapter)}
  & 75.35$\pm$0.71
  & 63.98$\pm$3.11
  & 48.74$\pm$0.62
  & 27.70$\pm$0.90
  & 53.94 \\

  F & Ditto
  & 74.56$\pm$0.31
  & \underline{65.45$\pm$4.25}
  & \underline{48.74$\pm$1.01}
  & \textbf{30.36$\pm$1.21}
  & \textbf{54.78} \\

  G & \method{} (Ours)
  & \textbf{76.20$\pm$0.48}
  & \textbf{66.10$\pm$7.85}
  & 47.40$\pm$1.32
  & 27.65$\pm$0.61
  & \underline{54.34} \\

  \bottomrule
  \end{tabular*}

  \par\medskip
  \noindent\textit{(b) Training protocols and nominal costs
  for federated configurations A--G}
  \par\smallskip

  \footnotesize

  \begin{tabularx}{\textwidth}
    {@{}c>{\raggedright\arraybackslash}Xccccc@{}}
  \toprule
  Code & Shared/private partition & $\mu$ or $\lambda$
  & \makecell{Client-held\\parameters\\(million)}
  & \makecell{Private\\parameters\\(\%)}
  & \makecell{Local epochs\\per round}
  & \makecell{Communication\\per round\\(MiB)} \\
  \midrule

  A & Full-model aggregation
    & $\mu{=}0$
    & 0.234--0.241 & 0 & 2 & 1.788--1.839 \\

  B & Full-model aggregation
    & $\mu{=}0.001$
    & 0.234--0.241 & 0 & 2 & 1.788--1.839 \\

  C & Encoder aggregated; classifier private
    & --
    & 0.230--0.237 & 1.87--1.98 & 2 & 1.721--1.770 \\

  D & Encoder aggregated; classifier private
    & $\mu{=}0$
    & 0.230--0.237 & 1.87--1.98 & 1+1 & 1.721--1.770 \\

  E & Encoder aggregated; classifier private
    & $\mu{=}0.0006$
    & 0.230--0.237 & 1.87--1.98 & 2 & 1.721--1.770 \\

  F & Global model $w$ aggregated;
      personalized model $v_k$ private
    & $\lambda{=}0.1$
    & 0.460--0.473 & 100 & 2+2 & 1.755--1.806 \\

  G & Encoder aggregated;
      adapter and classifier private
    & $\mu{=}0.0006$
    & 0.234--0.241 & 3.63--3.75 & 2 & 1.721--1.770 \\

  \bottomrule
  \end{tabularx}

  \par\smallskip
  \parbox{\linewidth}{\footnotesize
  Client-held parameters include all model copies but exclude
  optimizer states, gradients, and activations.
  Private parameter ratios are calculated relative to the
  inference-model parameter count (one base model for Ditto).
  Communication costs represent FP32 parameter downloads
  and uploads per client per round
  ($1\,\mathrm{MiB}$ (mebibyte) $=2^{20}$ bytes),
  excluding protocol overhead.
  For FedAvg and FedProx, the reported communication includes
  all parameters in the instantiated full-model state,
  including adapter tensors.
  In panel (a), bold and underlined values indicate the
  best and second-best reported means, respectively,
  among the federated configurations A--G.
}

\end{table*}

The 18 configurations combine different model families and training strategies. Section~\ref{sec:same-backbone} compares federated methods using the same hybrid backbone, and Section~\ref{sec:ablation} examines representation and personalization components under specified optimization settings.

\subsection{Same-Backbone FL Comparison}
\label{sec:same-backbone}

Table~\ref{tab:same-backbone} compares seven federated configurations using the same hybrid analog--spike backbone and evaluation procedure. Method-specific training schedules are retained: Ditto trains a global model $w$ and a personalized model $v_k$ with $\lambda=0.1$, while FedRep assigns one local epoch each to the private head and shared representation. Central and Local provide non-federated reference results.

With the same hybrid backbone, PFedHASNet obtains a four-dataset mean-client macro-F1 of 54.34\%, compared with 50.44\% for FedAvg, 50.52\% for FedProx, 54.07\% for FedPer, 54.03\% for FedRep, and 54.78\% for Ditto. PFedHASNet uses 0.234--0.241 million client-held parameters, approximately 49\% fewer than Ditto's 0.460--0.473 million, with a 0.44-pp lower mean macro-F1. Under the evaluated schedules, it also uses two local epochs per round versus Ditto's four nominal epochs. These results show the model-state and local-training costs of the two personalized strategies alongside their decoding performance.

\begin{figure*}[!t]
  \centering
  \includegraphics[width=0.9\textwidth]{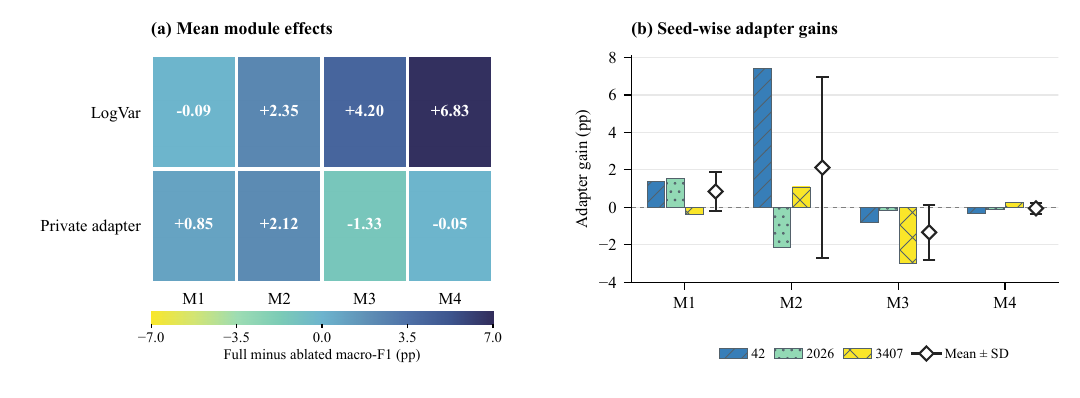}
  \caption{Module ablations at round 100 ($\mu=0.0006$).
  (a) Full-minus-ablated mean-client test macro-F1 (pp) on M1--M4.
  (b) Adapter gains for seeds 42, 2026, and 3407; diamonds and
  whiskers show mean $\pm$ sample SD.
  Ablations remove the LogVar branch or private residual adapter,
  respectively; the latter retains the private classifier.}
  \label{fig:module-signed-gains}
\end{figure*}

For the adapter-free configurations, comparisons
between $\mu=0$ and $\mu=0.0006$ yielded
dataset-dependent differences of $-0.10\pm0.08$,
$-0.40\pm0.33$, $+0.15\pm0.12$, and
$-0.15\pm0.20$ pp on M1--M4, respectively.
With $\mu$ fixed at $0.0006$, the full-minus-adapter-free
differences across seeds 42, 2026, and 3407 were
$+0.85\pm1.06$, $+2.12\pm4.85$,
$-1.33\pm1.48$, and $-0.05\pm0.29$ pp.
Figure~\ref{fig:module-signed-gains}(b) presents
these adapter-related differences separately for
each seed, together with their mean and sample SD.
The seed-wise results provide a more detailed account
of the variability underlying the dataset-level
comparisons.

Across M1--M4, PFedHASNet contains 225,553--232,051
shared parameters and 8,739--8,999 private parameters
per client. The private adapter and classifier account for
3.63\%--3.75\% of the inference-model parameters. Counting
one download and one upload of shared FP32 parameters per
client per round, the nominal communication payload is
\begin{equation}
C_{\mathrm{round}}=\frac{2\times4N_{\mathrm{shared}}}{2^{20}}
\quad\text{MiB},
\label{eq:communication-payload}
\end{equation}
where $N_{\mathrm{shared}}$ denotes the number of shared
parameters and $1\,\mathrm{MiB}=2^{20}$ bytes.
PFedHASNet exchanges 1.721--1.770 MiB per client per
round, approximately 2\% less than Ditto
(1.755--1.806 MiB). It also requires approximately
49\% fewer client-held parameters
(0.234--0.241 M versus 0.460--0.473 M).
PFedHASNet performs two local training epochs per
round, whereas Ditto trains its global and personalized
models for two epochs each.

\begin{table}[!t]
\centering
\caption{Client-level overall configuration differences: PFedHASNet minus Hybrid-FedAvg at round 100. Each value is in percentage points (pp).}
\label{tab:formal-client-gains}
\KBSColumnTableFit{%
\begin{tabular*}{\columnwidth}{@{\extracolsep{\fill}}lccc@{\extracolsep{0pt}\;to\;}c@{}}
\toprule
Dataset & Positive clients & Mean (pp) & \multicolumn{2}{c@{}}{Range (pp)} \\
\midrule
M1 & 3/9  & $+0.32$ & $-4.77$  & $+14.51$ \\
M2 & 5/7  & $+2.23$ & $-13.86$ & $+18.90$ \\
M3 & 7/9  & $+8.24$ & $-6.46$  & $+20.06$ \\
M4 & 9/10 & $+4.80$ & $-1.97$  & $+11.32$ \\
\bottomrule
\end{tabular*}}
\end{table}

\begin{figure*}[!tp]
  \centering
  \includegraphics[width=0.9\textwidth,keepaspectratio]{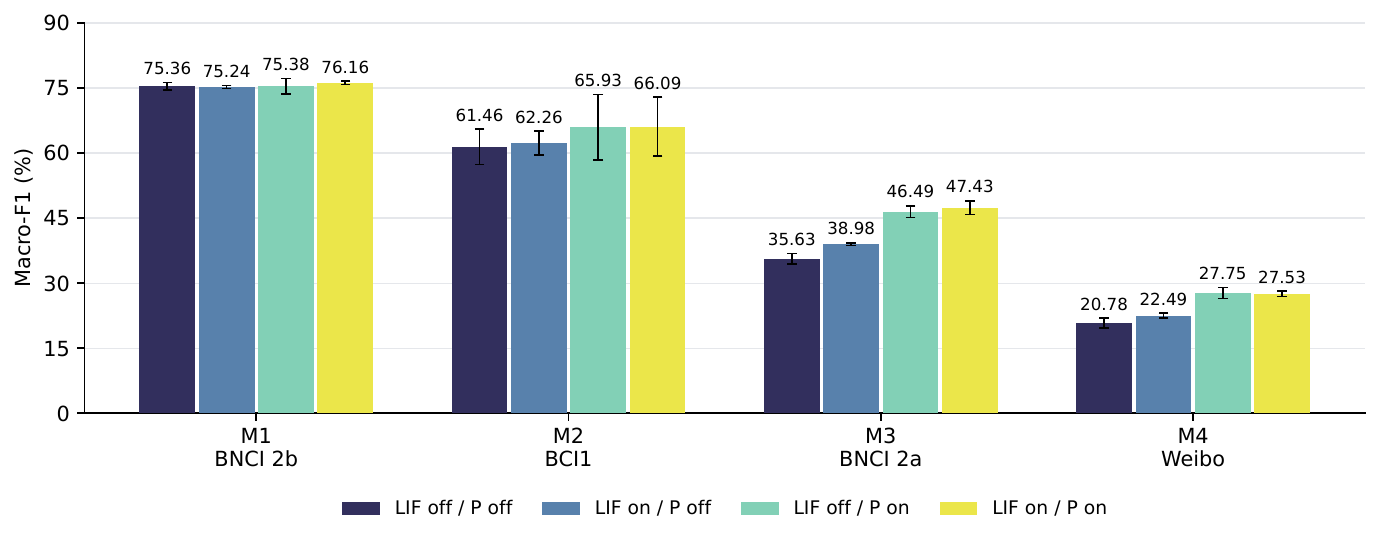}
  \caption{Matched $2\times2$ LIF--personalization ablation
  at round 100 ($\mu=0$), with the analog and LogVar
  branches retained. Bars show mean-client test macro-F1
  (\%) across three seeds, with sample SD error bars.
  The four LIF--personalization configurations are
  compared on M1--M4. Conditional gains and interactions
  are reported in
  Table~\ref{tab:lif-personalization-interaction}.}
  \label{fig:lif-personalization-interaction}
\end{figure*}

Table~\ref{tab:formal-client-gains} reports client-level
test macro-F1 differences at round 100 between PFedHASNet
and Hybrid-FedAvg across 35 participating clients on M1--M4.
For each client, the difference is calculated within each
of the matched random seeds 42, 2026, and 3407 and then
averaged across seeds. \emph{Positive clients} counts
clients whose three-seed mean difference is
greater than zero; \emph{Mean} averages these client
differences with equal weight, while \emph{Range} denotes
the minimum and maximum client means. All differences
are expressed in percentage points (pp), and the
ranges represent observed client-level extremes rather
than standard deviations or confidence intervals.

\begin{table*}[!t]
  \centering
  \caption{Conditional personalization and LIF gains in the matched
  $2\times2$ factorial ablation at round 100 ($\mu=0$).
  Values are within-seed paired mean $\pm$ sample SD across
  seeds 42, 2026, and 3407 (pp). The interaction $I$ is the
  common difference of differences for the two factors.}
  \label{tab:lif-personalization-interaction}
  \small\normalfont
  \setlength{\tabcolsep}{5pt}
  \renewcommand{\arraystretch}{1.18}
  \begin{tabular*}{\textwidth}{@{\extracolsep{\fill}}lccccc@{}}
    \toprule
    & \multicolumn{2}{c}{Personalization gain (pp)}
    & \multicolumn{2}{c}{LIF gain (pp)}
    & \multicolumn{1}{c}{Interaction (pp)} \\
    \cmidrule(lr){2-3}\cmidrule(lr){4-5}\cmidrule(l){6-6}
    Dataset & LIF off & LIF on & P off & P on & $I$ \\
    \midrule
    M1 & $+0.02\pm0.92$ & $+0.91\pm0.19$ & $-0.11\pm0.61$ & $+0.78\pm1.43$ & $+0.89\pm0.84$ \\
    M2 & $+4.47\pm10.31$ & $+3.83\pm6.35$ & $+0.80\pm2.76$ & $+0.16\pm1.50$ & $-0.64\pm4.06$ \\
    M3 & $+10.86\pm0.94$ & $+8.45\pm1.90$ & $+3.36\pm1.38$ & $+0.95\pm0.38$ & $-2.41\pm1.42$ \\
    M4 & $+6.97\pm0.53$ & $+5.04\pm0.35$ & $+1.71\pm1.02$ & $-0.22\pm0.64$ & $-1.93\pm0.79$ \\
    \bottomrule
  \end{tabular*}
  \par\smallskip
  \parbox{\textwidth}{\footnotesize
  P denotes personalization. Each conditional gain is calculated
  within seed before averaging; its sample SD is computed from
  the three paired differences. Analog and LogVar remain enabled.
  The $I$ column is reported once because both conditional
  comparisons yield the same factorial interaction.}
\end{table*}

\begin{figure*}[!tp]
  \centering
  \includegraphics[width=0.96\textwidth]{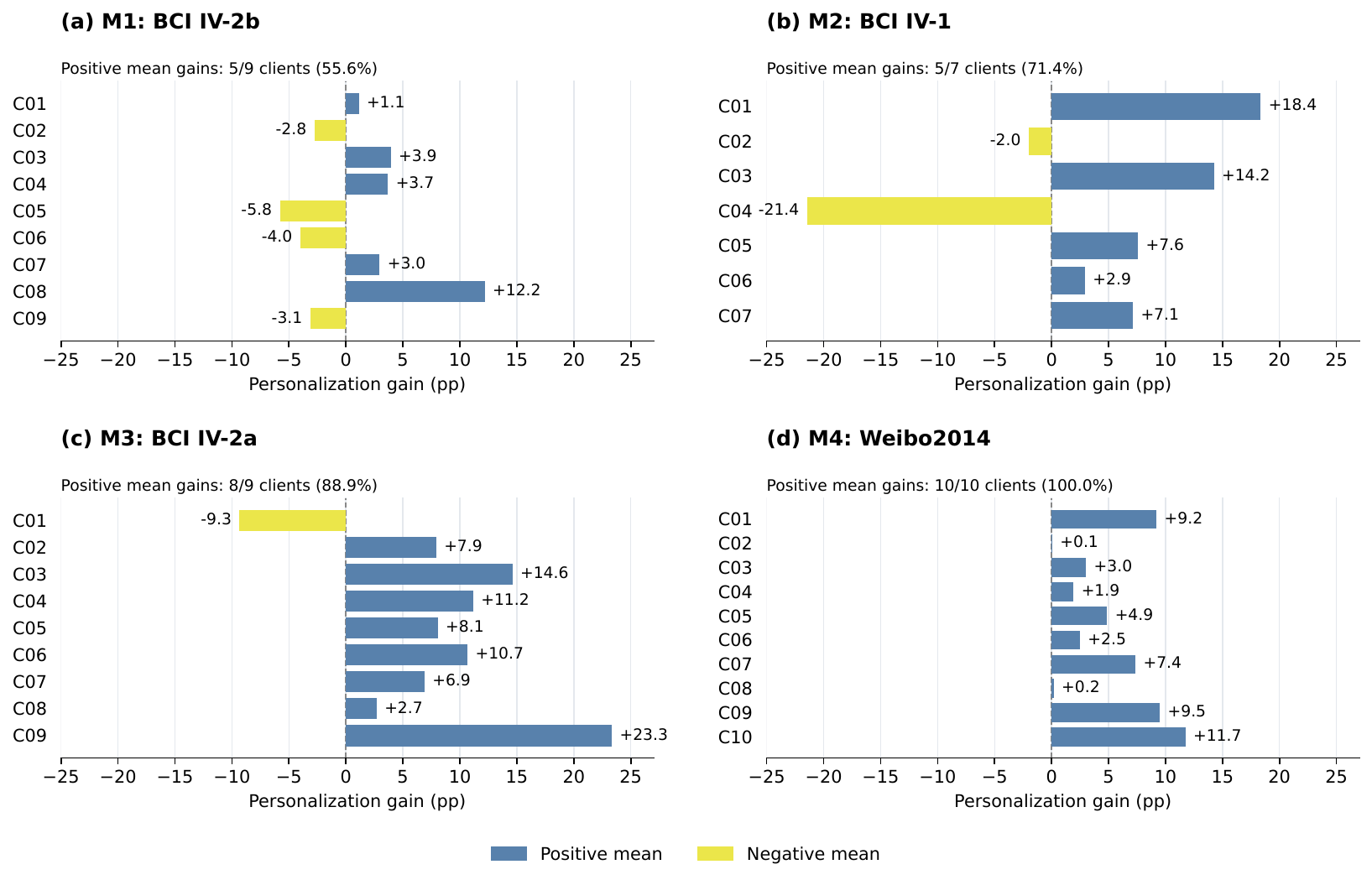}
  \caption{Client-level personalization gains at round 100
  (LIF on, $\mu=0$). Bars show mean paired personalization on--off
  differences in test macro-F1 across seeds 42, 2026, and 3407 (pp);
  whiskers indicate seed-wise minima and maxima, not SDs or confidence
  intervals. Blue and yellow denote positive and negative mean
  gains, respectively.}
  \label{fig:participating-client-gains}
\end{figure*}

All four datasets show positive mean differences
despite heterogeneous client-level effects.
M1 averages $+0.32$ pp (3/9 positive clients;
range $-4.77$ to $+14.51$ pp), M2 $+2.23$ pp
(5/7; $-13.86$ to $+18.90$ pp), M3 $+8.24$ pp
(7/9; $-6.46$ to $+20.06$ pp), and M4 $+4.80$ pp
(9/10; $-1.97$ to $+11.32$ pp).
Overall, 24/35 clients exhibit positive three-seed
mean differences, although negative outcomes occur
across all datasets.

These results compare the complete PFedHASNet
configuration ($\mu=0.0006$) with Hybrid-FedAvg
($\mu=0$), which also differs in client-private
personalization and shared--private parameter partitioning.
The reported differences therefore characterize the
complete configurations without isolating the contribution
of any individual component. The client-level mean differences
and their ranges are summarized in
Table~\ref{tab:formal-client-gains}.
Figure~\ref{fig:participating-client-gains} presents
a separate matched personalization on--off comparison
with LIF enabled at $\mu=0$. For example, the matched
comparison records positive mean gains for 5/9 clients
on M1, compared with 3/9 under the complete-configuration
comparison in Table~\ref{tab:formal-client-gains}.

\begin{figure*}[!tp]
  \centering
  \includegraphics[width=0.96\textwidth,keepaspectratio]
  {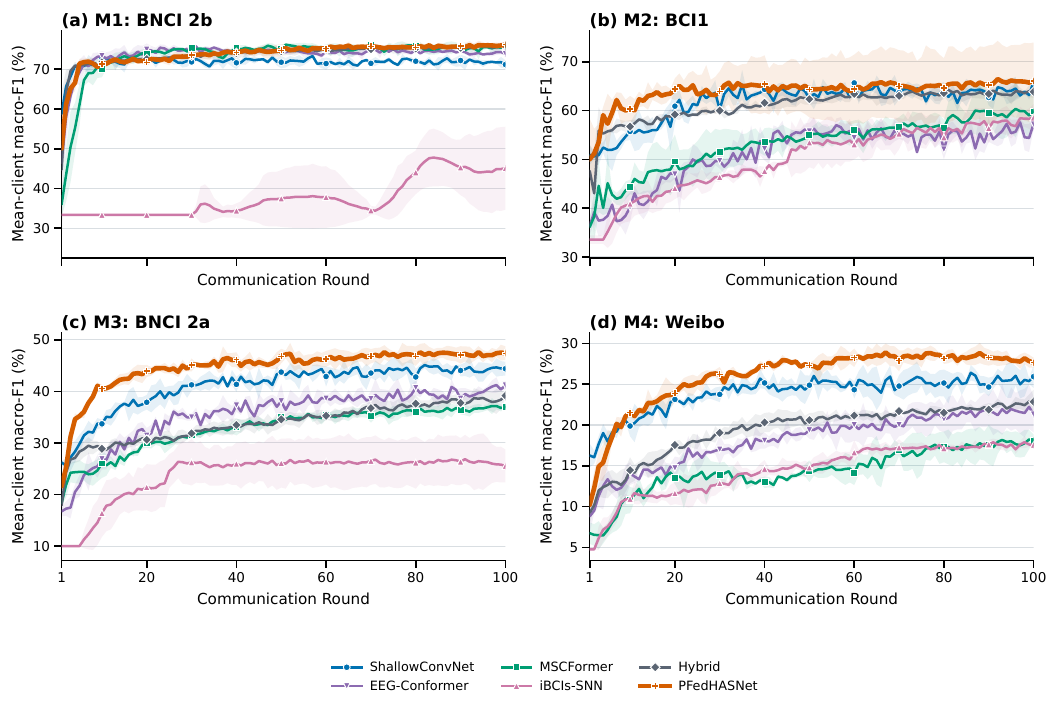}
  \caption{Post-aggregation mean-client test macro-F1 (\%) on M1--M4 for six configurations. Curves and shading show mean $\pm$ sample SD across three seeds; enlarged markers indicate round 100. Hybrid denotes Hybrid-FedAvg.}
  \label{fig:representative-convergence}
\end{figure*}

\subsection{Representation and Personalization Ablations}
\label{sec:ablation}

Figure~\ref{fig:module-signed-gains}(a) evaluates the
contributions of channel-wise LogVar statistics in HASR
and the client-private residual adapter at round 100
($\mu=0.0006$). The LogVar branch improves mean-client
macro-F1 by $+2.35$, $+4.20$, and $+6.83$ pp on M2--M4,
respectively, with a slight decrease of $0.09$ pp on M1.
The average gain of $3.32$ pp across four datasets
indicates that channel-wise amplitude statistics
complement the learned temporal and spike-rate features,
particularly on M3 and M4.

The private residual adapter provides additional
subject-specific feature correction, yielding gains of
$+0.85$ and $+2.12$ pp on M1 and M2, respectively.
Its effects are smaller or negative on M3 ($-1.33$ pp)
and M4 ($-0.05$ pp). The adapter-free configuration
retains the private classifier, isolating the contribution
of residual feature correction. Figure~\ref{fig:module-signed-gains}(b)
further shows substantial seed-wise variation on M2
($+2.12\pm4.85$ pp) and negative differences across
all three seeds on M3. These results highlight the
contribution of channel-wise statistics to hybrid EEG
representation and the dataset-dependent benefits of
client-private adaptation. At the same proximal coefficient
and reporting round, the four-dataset average macro-F1 is
54.34\% for PFedHASNet and 51.02\% without LogVar; the
per-dataset signed differences are shown in
Figure~\ref{fig:module-signed-gains}(a).

To examine how the LIF spike-rate pathway in HASR
interacts with client personalization, we conduct a
matched $2\times2$ factorial ablation at round 100
($\mu=0$), retaining the continuous-valued and LogVar
branches. The four configurations are evaluated on
M1--M4 using three matched random seeds.
As shown in Figure~\ref{fig:lif-personalization-interaction}
and Table~\ref{tab:lif-personalization-interaction},
personalization yields larger average macro-F1 gains
than LIF, while the contribution of LIF varies across
datasets and personalization states. These results
highlight the role of client-specific adaptation and
the conditional contribution of spike-rate features
within the hybrid representation. The absolute condition
means are labeled in Figure~\ref{fig:lif-personalization-interaction},
while the within-seed paired gains, sample SDs, and
interactions appear in
Table~\ref{tab:lif-personalization-interaction}.

As shown in
Figure~\ref{fig:lif-personalization-interaction},
the configuration with both LIF and personalization
enabled achieves the highest four-dataset average
mean-client macro-F1 of 54.30\%, exceeding the
configuration with both components disabled by
5.99 percentage points (pp).
Across M1--M4, the personalization-on bars
exceed their corresponding personalization-off
bars under both LIF states, although the
differences vary across datasets.
The seed-paired comparisons in
Table~\ref{tab:lif-personalization-interaction}
yield four-dataset average personalization gains
of +5.58 pp with LIF disabled and +4.56 pp
with LIF enabled, using equal dataset weights.
Both gains exceed the corresponding average
conditional LIF gains at fixed personalization
states.

\begin{figure*}[!t]
  \centering
  \includegraphics[width=0.96\textwidth,keepaspectratio]
  {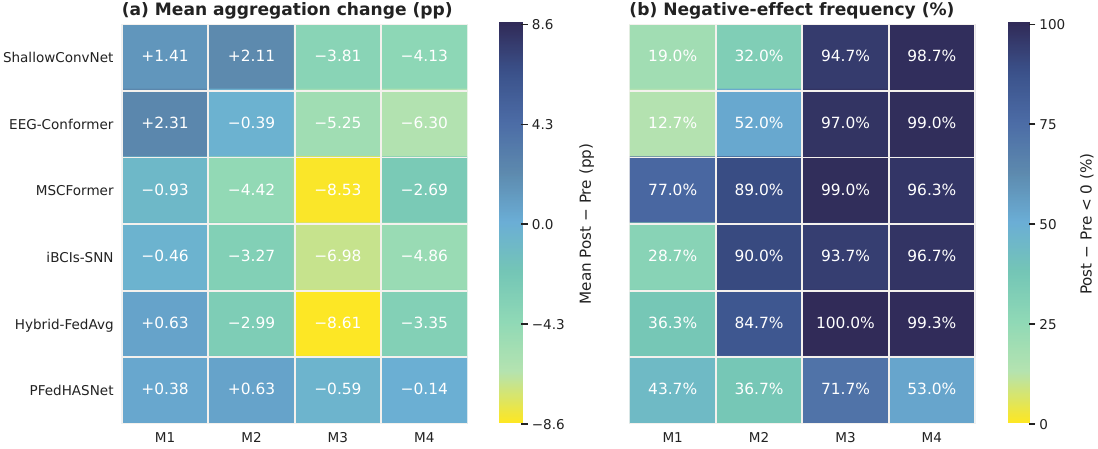}
  \caption{Within-round aggregation effects on mean-client test macro-F1. (a) Mean post$-$pre change (MeanAgg, pp), with a zero-centered scale. (b) Frequency of strictly negative changes (\%; scale 0--100). Each cell summarizes three seeds $\times$ 100 rounds.}
  \label{fig:aggregation-effect-summary}
\end{figure*}

The dataset-level comparisons further reveal
differences in the magnitude and seed-wise
variability of personalization gains. On M3,
personalization improves mean-client macro-F1
by $10.86\pm0.94$ pp with LIF disabled and
$8.45\pm1.90$ pp with LIF enabled.
The corresponding gains on M4 are
$6.97\pm0.53$ and $5.04\pm0.35$ pp,
respectively. In comparison, M1 shows
a near-zero mean gain without LIF
($+0.02\pm0.92$ pp), increasing to
$+0.91\pm0.19$ pp when LIF is enabled.
M2 also records positive mean gains under
both conditions (+4.47 and +3.83 pp),
although their paired sample SDs are
considerably larger (10.31 and 6.35 pp).
Although the magnitude and seed-wise variability of
personalization gains differ considerably across datasets,
positive mean personalization gains are observed on all
four datasets under both LIF states.

Table~\ref{tab:lif-personalization-interaction} also reports
conditional LIF gains with personalization held fixed,
alongside the conditional personalization gains.
On M1, the mean LIF gain changes from $-0.11$ pp with
personalization off to $+0.78$ pp with personalization on.
On M2, the corresponding gains are $+0.80$ and
$+0.16$ pp, accompanied by paired sample SDs of
2.76 and 1.50 pp. On M3 and M4, positive LIF gains
without personalization decrease after personalization
is enabled: from $+3.36$ to $+0.95$ pp on M3,
and from $+1.71$ to $-0.22$ pp on M4.
The shared interaction $I$ is positive on M1
($+0.89$ pp) and negative on M2--M4
($-0.64$, $-2.41$, and $-1.93$ pp, respectively).
These signed contrasts describe changes in the
conditional effects of one factor across the two
states of the other factor.

We next examine participating-client personalization in the matched LIF-on setting. Figure~\ref{fig:participating-client-gains} shows positive mean paired gains for 5/9, 5/7, 8/9, and 10/10 clients on M1--M4, respectively. These client-level profiles provide a complementary view of the heterogeneous personalization effects summarized by dataset-level means.

Let $F_{lp}^{(s)}$ denote the round-100 mean-client macro-F1
(\%) for seed $s$, with $l$ and $p$ indicating the LIF
and personalization states (0: off; 1: on). Within each
seed, the personalization gains are
$F_{01}^{(s)}-F_{00}^{(s)}$ and
$F_{11}^{(s)}-F_{10}^{(s)}$, while the LIF gains are
$F_{10}^{(s)}-F_{00}^{(s)}$ and
$F_{11}^{(s)}-F_{01}^{(s)}$. Their interaction is
$I^{(s)}=(F_{11}^{(s)}-F_{10}^{(s)})-
(F_{01}^{(s)}-F_{00}^{(s)})$.
Table~\ref{tab:lif-personalization-interaction} reports
the mean and sample SD of these paired differences across
seeds 42, 2026, and 3407. The mean interactions are
$+0.89$, $-0.64$, $-2.41$, and $-1.93$ pp on M1--M4,
respectively. Personalization yields positive mean gains
under both LIF settings across all four datasets, with
larger improvements on M3 and M4. These results highlight
the contribution of client-specific adaptation to MI-EEG
decoding, while the LIF pathway provides additional
dataset-dependent effects.

The LIF pathway introduces membrane integration,
temporal leakage, threshold-triggered firing, and
reset dynamics, providing a spike-based temporal
representation alongside the continuous-valued and
statistical branches. It adds only 130 trainable
parameters (approximately 0.054\%--0.057\% of the
configuration without LIF) and yields a positive
four-dataset mean conditional gain with personalization
enabled. Its lightweight spiking dynamics also provide
a basis for exploring future neuromorphic edge deployment.


\subsection{Training Dynamics and Aggregation Effects}
\label{sec:training-dynamics}

We compare ShallowConvNet, EEG-Conformer, MSCFormer, iBCIs-SNN, Hybrid-FedAvg, and PFedHASNet on held-out test trials of participating subjects. The first five configurations use FedAvg; Hybrid-FedAvg and PFedHASNet share the hybrid backbone but use different personalization and proximal-regularization settings ($\mu=0$ and $\mu=0.0006$). Figure~\ref{fig:representative-convergence} presents the post-aggregation trajectories, whereas Figure~\ref{fig:aggregation-effect-summary} reports the paired performance changes immediately before and after each aggregation step. The shaded intervals in Figure~\ref{fig:representative-convergence}
show between-seed variation in post-aggregation performance.
The within-round differences and their negative frequencies
are summarized separately in
Figure~\ref{fig:aggregation-effect-summary}.

For a fixed configuration and dataset, let
$F_{\mathrm{Pre}}^{(s)}(t)$ and $F_{\mathrm{Post}}^{(s)}(t)$ denote
fraction-valued mean-client test macro-F1 on the same client test
splits, immediately before and after server aggregation, respectively.Pre is evaluated after the current round's local updates; it is not the independently trained Local baseline in
Table~\ref{tab:same-backbone}. For \method{}, Post combines the
aggregated shared encoder with the client-private
modules. The aggregation-induced change in percentage points is
\begin{equation}
\Delta_{\mathrm{Agg}}^{(s)}(t)
=100\left[F_{\mathrm{Post}}^{(s)}(t)
-F_{\mathrm{Pre}}^{(s)}(t)\right].
\label{eq:aggregation-change}
\end{equation}
With $\mathcal{S}=\{42,2026,3407\}$ and $T=100$, we calculate
\begin{equation}
\mathrm{MeanAgg}
=\frac{1}{|\mathcal{S}|T}
\sum_{s\in\mathcal{S}}\sum_{t=1}^{T}
\Delta_{\mathrm{Agg}}^{(s)}(t)
\label{eq:mean-aggregation-change}
\end{equation}
and the negative-effect frequency (\%) as
\begin{equation}
p_{\mathrm{neg}}
=\frac{100}{|\mathcal{S}|T}
\sum_{s\in\mathcal{S}}\sum_{t=1}^{T}
\mathbf{1}\!\left[\Delta_{\mathrm{Agg}}^{(s)}(t)<0\right].
\label{eq:negative-aggregation-frequency}
\end{equation}
Each configuration--dataset summary contains 300 paired seed--round observations. Negative-change frequency counts changes below zero before averaging across seeds; it is a frequency over seed--round observations, not a proportion of clients. Because rounds within a run are temporally dependent, these values are reported descriptively without treating the 300 observations as independent samples for significance testing.

Figure~\ref{fig:aggregation-effect-summary} reports the
mean post-minus-pre aggregation change (MeanAgg) and
the frequency of strictly negative changes across the
recorded seed--round observations. For \method{},
MeanAgg is $+0.38$ and $+0.63$ pp on M1 and M2,
respectively, while the corresponding negative-change
frequencies are 43.7\% and 36.7\%. On M3 and M4,
the mean changes are $-0.59$ and $-0.14$ pp, with
negative-change frequencies of 71.7\% and 53.0\%,
respectively. The positive mean changes on M1 and M2
coexist with negative changes in a subset of rounds
and seeds, while M3 and M4 exhibit negative mean
changes with different frequencies of negative
observations.

The comparison with Hybrid--FedAvg varies across
datasets and across the two reported measures.
On M1, \method{} records a smaller positive MeanAgg
than Hybrid--FedAvg ($+0.38$ versus $+0.63$ pp),
together with a higher negative-change frequency
(43.7\% versus 36.3\%). On M2, the difference is
reversed: \method{} has a positive mean change of
$+0.63$ pp compared with $-2.99$ pp for
Hybrid--FedAvg, and its negative-change frequency
is 36.7\% rather than 84.7\%. The differences are
also pronounced on M3 and M4. On M3, the mean
changes are $-0.59$ and $-8.61$ pp, with
negative-change frequencies of 71.7\% and
100.0\% for \method{} and Hybrid--FedAvg,
respectively. On M4, the corresponding values
are $-0.14$ versus $-3.35$ pp and 53.0\%
versus 99.3\%. Across M2--M4, \method{} records
higher MeanAgg values and lower negative-change
frequencies than Hybrid--FedAvg, whereas the
latter has more favorable values on both measures
for M1.

Across M2--M4, PFedHASNet records higher mean post-minus-pre aggregation changes and fewer negative-change observations than Hybrid-FedAvg; M1 shows the reverse pattern. These results describe the complete training configurations, which differ in private parameters and proximal regularization. Figure~\ref{fig:representative-convergence} provides post-aggregation performance trajectories.


\section{Conclusion and Future Work}
\label{sec:conclusion}

PFedHASNet integrates shared hybrid representation learning with client-private adaptation to enable collaborative MI-EEG decoding while keeping raw data local. Experiments on four public datasets demonstrate that the framework achieves competitive decoding performance on held-out trials from participating subjects while maintaining a compact client-side model. Ablation analyses indicate that personalization and channel-wise log-variance statistics are major contributors to performance improvements, whereas the benefits of the spike branch and residual adapters vary across datasets and configurations. These findings support retaining subject-specific feature refinement and prediction modules alongside shared EEG representations, providing a feasible approach to personalized brain–computer interface modeling with decentralized data. Future work will investigate adaptation across recording sessions and explore personalized decoding for previously unseen subjects using limited calibration data.

\section*{References}
\bibliographystyle{IEEEtran}
\bibliography{references}

\end{document}